\pdfoutput=1

\documentclass[11pt]{article}

\usepackage[final]{coling}

\usepackage{times}
\usepackage{latexsym}
\usepackage[T1]{fontenc}

\usepackage[utf8]{inputenc}

\usepackage{inconsolata}

\usepackage{graphicx}
\usepackage{booktabs} 
\usepackage{mathabx}
\usepackage{multirow}
\usepackage{setspace}
\usepackage[none]{hyphenat}
\usepackage{xspace}
\usepackage{tabularx}
\usepackage{tabu}
\usepackage[english]{babel}
\usepackage{microtype}
\usepackage{lipsum}
\usepackage{hyperref}
\usepackage{tabularray}
\usepackage{graphics}
\usepackage{float}
\title{CamemBERT 2.0: A Smarter French Language Model Aged to Perfection}

\author{Wissam Antoun \quad Francis Kulumba \quad Rian Touchent \\
\large\textbf{\quad Éric de la Clergerie \quad Benoît Sagot \quad Djamé Seddah} \\
     Inria, Paris, France \\
     \{firstname,lastname\}@inria.fr}

\begin{document}
\maketitle

\begin{abstract}
French language models, such as CamemBERT, have been widely adopted across industries for natural language processing (NLP) tasks, with models like CamemBERT seeing over 4 million downloads per month.
However, these models face challenges due to temporal concept drift, where outdated training data leads to a decline in performance, especially when encountering new topics and terminology. 
This issue emphasizes the need for updated models that reflect current linguistic trends.
In this paper, we introduce two new versions of the CamemBERT base model—CamemBERTav2 and CamemBERTv2—designed to address these challenges. 
CamemBERTav2 is based on the DeBERTaV3 architecture and makes use of the Replaced Token Detection (RTD) objective for better contextual understanding, while CamemBERTv2 is built on RoBERTa, which uses the Masked Language Modeling (MLM) objective. 
Both models are trained on a significantly larger and more recent dataset with longer context length and an updated tokenizer that enhances tokenization performance for French.
We evaluate the performance of these models on both general-domain NLP tasks and domain-specific applications, such as medical field tasks, demonstrating their versatility and effectiveness across a range of use cases. 
Our results show that these updated models vastly outperform their predecessors, making them valuable tools for modern NLP systems.
All our new models, as well as intermediate checkpoints, are made openly available on Huggingface \url{https://huggingface.co/almanach?search_models=camembert+v2}.
\end{abstract}

\section{Introduction}
In recent years, French language models such as CamemBERT~\cite{martin-etal-2020-camembert} have become integral to businesses leveraging natural language processing (NLP) to boost productivity and efficiency.
Since its release, CamemBERT has gained widespread adoption, receiving over 4 million downloads each month, and continues to be actively used by the NLP community.
A notable example is ENEDIS, which used CamemBERT to automate the dispatch of 100,000 customer requests per day across 1,500 operators, significantly reducing manual workload and achieving a return on investment of approximately €3M per year~\cite{Gemignani2023enedis,Akani2023EneBERT}.

However, a significant challenge that models like CamemBERT, developed in 2019, face is temporal concept drift~\cite{loureiro-etal-2022-timelms,agarwal-nenkova-2022-temporal,jin-etal-2022-lifelong}
This phenomenon occurs when the data a model was originally trained on becomes outdated, leading to a decline in performance as new topics, events, and terminology emerge. 
For instance, when CamemBERT was trained, discussions around COVID-19, public health restrictions, and associated changes in language usage were not present in the training data. As a result, models like CamemBERT, which have not been updated, struggle to understand or generate accurate responses to these newer concepts. 
Temporal drift impacts their ability to remain relevant in evolving real-world applications, highlighting the need for continuous updates to keep the models aligned with current linguistic and contextual trends.

Given these challenges, it is essential to develop and deploy updated encoder models that can better serve modern NLP applications. 
In response, we aim to provide state-of-the-art models and fine-tuned versions for a range of common NLP tasks, including Named Entity Recognition (NER), Question Answering (QA), Natural Language Inference (NLI), and Part-of-Speech (POS) tagging. 
These updated models will play a major role in creating fast, efficient, and reliable AI systems.

In this paper, we present two new versions of the CamemBERT base model: CamemBERTav2 and CamemBERTv2. 
CamemBERTav2 is built on the DeBERTaV3~\cite{he2021debertav3} architecture as an update to the CamemBERTa model~\cite{antoun2023data}, using the Replaced Token Detection (RTD) training objective for enhanced context and positional representation, while CamemBERTv2 is based on the RoBERTa~\cite{liu2019roberta} architecture, trained using the Masked Language Modeling (MLM) objective. 
Both models benefit from training on a much larger and more recent dataset, coupled with an updated tokenizer designed to better capture the nuances of the French language, and to support modern token requirements by adding new lines, tabulation and emojies to the vocabulary.
This ensures improved tokenization performance across various NLP tasks.

To evaluate the performance of these new models, we conducted extensive tests on both general-domain NLP tasks and domain-specific applications, such as those in the medical field. 
This dual approach demonstrates the versatility of our models, highlighting their capacity to excel in diverse use cases, including highly specialized areas where domain-specific knowledge is necessary.

The contributions of this paper are as follows:
\begin{itemize}
    \item We present CamemBERTav2 and CamemBERTv2 and trained on a larger, up-to-date dataset with an enhanced tokenizer to better capture the complexities of the French language.
    \item We evaluate the models on both general-domain NLP tasks and domain-specific applications, particularly in the medical field, to demonstrate their robustness and versatility.
    \item We are releasing all model artifacts, including intermediate checkpoints and fine-tuned models, enabling the community to further experiment, fine-tune, and deploy these models across various applications.\footnote{\url{https://huggingface.co/almanach?search_models=camembert+v2}}
\end{itemize}
These contributions aim to advance French language modeling and provide the community with state-of-the-art tools for diverse NLP tasks.
\section{Related Works}
\paragraph{Pre-trained French Language Models}
French language models have predominantly been trained using either Masked Language Modeling (MLM) or Causal Language Modeling (CLM) techniques. 
Among the leading French models, CamemBERT~\cite{martin-etal-2020-camembert} and FlauBERT~\cite{le-etal-2020-flaubert-unsupervised} have gained considerable traction, both relying on the MLM pre-training method.
Other noteworthy models include FrALBERT~ \cite{cattan:hal-03336060}, a French adaptation of ALBERT~\cite{Lan2020ALBERT}, and LePetit~\cite{micheli-etal-2020-importance}, a scaled-down version of CamemBERT. In addition, D’AlemBERT~\cite{gabay-etal-2022-freem}, derived from RoBERTa~\cite{liu2020roberta}, is tailored for Early Modern French. BARThez~\cite{kamal-eddine-etal-2021-barthez} is a sequence-to-sequence model built on BART’s architecture~\cite{lewis-etal-2020-bart}, while PAGnol~\cite{launay-etal-2022-pagnol} and Cedille~\cite{muller2022cedille} represent models trained with CLM.

Recently, CamemBERTa, a French DeBERTa model based on the DeBERTaV3 architecture, has been introduced. 
CamemBERTa demonstrates superior performance on various French NLP tasks compared to traditional BERT-based models, despite using significantly fewer training tokens~\cite{antoun2023data}. 
Additionally, CamemBERT-bio, a specialized model fine-tuned for French biomedical data, has shown substantial improvements in named entity recognition tasks within the biomedical domain, addressing the limitations of general-purpose models like CamemBERT when applied to specialized texts~\cite{touchent-de-la-clergerie-2024-camembert}. 
DrBERT further extends this focus by pre-training on both public and private medical data, offering specialized performance for the French biomedical field~\cite{labrak-etal-2023-drbert}.
Finally, CharacterBERT-French, a character-based variant of BERT, offers robustness to noise by utilizing character embeddings instead of subword-based vocabularies. This model has shown promise, particularly in handling noisy experimental data~\cite{riabi-etal-2021-character}.
\section{CamemBERT 2.0}
In this paper, we introduce two updated versions of the CamemBERT base model, named CamemBERTav2 and CamemBERTv2, developed to improve upon the original CamemBERTa~\cite{antoun2023data} and CamemBERT~\cite{martin-etal-2020-camembert} models respectively.
CamemBERTav2 is built on the DeBERTaV3~\cite{he2021debertav3,he2021deberta} architecture and the Replaced Token Detection (RTD)~\cite{clark2020electra} training objective, leveraging its improved attention mechanism for better context and positional representation. 
CamemBERTv2, on the other hand, is based on RoBERTa~\cite{liu2019roberta}, using the Masked Language Modeling (MLM) objective for training, and is meant to be a drop-in replacement in task where computing having the pseudo-language modeling probability is needed.
The models were trained on a much larger and up-to-date dataset, accompanied by an updated tokenizer that better captures the linguistic complexities of the French language, ensuring improved tokenization performance for various downstream tasks.
\begin{table*}[!ht]
	
\centering
{\footnotesize
    \resizebox{\textwidth}{!}{%
        \begin{tabu}{ l  c  c @{\hspace{0.35cm}}  @{\hspace{0.35cm}} c  c @{\hspace{0.35cm}}  @{\hspace{0.35cm}} c  c  @{\hspace{0.35cm}}  @{\hspace{0.35cm}} c  c |c}
            \toprule
            & \multicolumn{2}{c @{\hspace{0.5cm}}}{\textsc{GSD}} & \multicolumn{2}{c @{\hspace{0.7cm}}}{\textsc{Rhapsodie}} & \multicolumn{2}{c @{\hspace{0.7cm}}}{\textsc{Sequoia}} & \multicolumn{2}{c @{\hspace{0.35cm}}}{\textsc{FSMB}} & {\sc FTB-NER} \\
            \cmidrule(l{2pt}r{0.4cm}){2-3}\cmidrule(l{-0.2cm}r{0.4cm}){4-5}\cmidrule(l{-0.2cm}r{0.4cm}){6-7}\cmidrule(l{-0.2cm}r{2pt}){8-9}\cmidrule(l{0cm}r{1pt}){10-10}
            \multirow{-2}{*}[1pt]{\textsc{Model}} & \textsc{UPOS}                        & \textsc{LAS}                         & \textsc{UPOS}                        & \textsc{LAS}                         & \textsc{UPOS}                        & \textsc{LAS}                         & \textsc{UPOS}                        & \textsc{LAS}                         & {\sc F1}                             \\
            \midrule
                        
            CamemBERT                   & 98.57{\scriptsize$\pm$0.07} & 94.35{\scriptsize$\pm$0.15} & 97.62{\scriptsize$\pm$0.08} & 84.29{\scriptsize$\pm$0.56} & 99.35{\scriptsize$\pm$0.09} & 94.78{\scriptsize$\pm$0.12} & 94.80{\scriptsize$\pm$0.16} & 81.34{\scriptsize$\pm$0.63} & 89.97{\scriptsize$\pm$0.50}          \\ %
            CamemBERTa                          & 98.55{\scriptsize$\pm$0.05} & 94.38{\scriptsize$\pm$0.15} & 97.52{\scriptsize$\pm$0.14} & 84.23{\scriptsize$\pm$0.08} & 99.44{\scriptsize$\pm$0.07} & 94.85{\scriptsize$\pm$0.14} & 94.80{\scriptsize$\pm$0.09} & 80.74{\scriptsize$\pm$0.25}          & 90.33{\scriptsize$\pm$0.54} \\
            \midrule
            CamemBERTv2                          & 98.60{\scriptsize$\pm$0.05} & 94.18{\scriptsize$\pm$0.12} & 97.62{\scriptsize$\pm$0.10} & 84.09{\scriptsize$\pm$0.31} & 99.37{\scriptsize$\pm$0.04} & 94.80{\scriptsize$\pm$0.14} & 95.05{\scriptsize$\pm$0.18} & 81.49{\scriptsize$\pm$0.38}          & 91.99{\scriptsize$\pm$0.96} \\
            CamemBERTav2                          & 98.54{\scriptsize$\pm$0.03} & 94.35{\scriptsize$\pm$0.20} & 97.70{\scriptsize$\pm$0.21} & 84.30{\scriptsize$\pm$0.27} & 99.42{\scriptsize$\pm$0.05} & 94.61{\scriptsize$\pm$0.25} & 95.19{\scriptsize$\pm$0.11} & 81.32{\scriptsize$\pm$0.29}          & \textbf{93.40}{\scriptsize$\pm$0.62} \\
            \bottomrule
        \end{tabu}
    }
}
\caption{\textbf{POS tagging}, \textbf{dependency parsing} and {\bf NER} results on the test sets of our French datasets. \textit{UPOS (Universal Part-of-Speech) refers here to POS tagging accuracy, and LAS measures the overall accuracy of labeled dependencies in a parsed sentence.} }
\label{tab:pos_and_dp_results}
\end{table*}

\subsection{Pre-Training Dataset}
Our new pre-training dataset is sourced mainly from the French subset of the CulturaX corpus~\cite{nguyen2023culturax}.
CulturaX is a multilingual dataset that combines mC4~\cite{xue-etal-2021-mt5} and four OSCAR~\cite{OrtizSuarezSagotRomary2019,AbadjiOrtizSuarezRomaryetal.2021,abadji-etal-2022-towards} snapshots.\footnote{Culturax contains the following OSCAR corpora 20.19, 21.09, 22.01, and 23.01, and the version 3.1.0 of mC4}
The documents are then deduplicated on the document level and filtered using language filters, URL block lists, and a comprehensive set of metric-based filters (e.g. stopword ratio, perplexity score, word repetition ratio...).
In addition, we make use of the French section of Wikipedia\footnote{We use the April 2024 dump}, and French scientific papers and theses from the HALvesting corpus~\cite{kulumba2024harvestingtextualstructureddata}.
In total, we gather 265B tokens from Culturax, 4.7B tokens from HALvesting, and 0.5B tokens from Wikipedia.
During training, we upsample Wikipedia 10 times, and hence our final pre-training dataset has 275B tokens, compared to 32B which were used during the original CamemBERT and CamemBERTa training.

\subsection{Tokenizer}
A key improvement in the CamemBERTv2 models is the development of an updated tokenizer. 
The primary goal was to improve tokenization efficiency by addressing the limitations of the previous version.
This includes the introduction of newline and tab characters, as well as support for emojis, which are normalized by removing zero-width joiner characters and splitting emoji sequences into individual tokens. 
To improve the handling of numerical data, we opted to split numbers into a maximum of two-digit tokens, which we hypothesize will enhance the model's ability to process dates and perform simple arithmetic tasks—functions more commonly utilized in encoder models than in generative ones. 
Additionally, French and English elisions (e.g., l', lorsqu') are now treated as single tokens, including the apostrophe. 
We adopted the WordPiece tokenization algorithm~\cite{devlin-etal-2019-bert}, which allows for flexible vocabulary adjustments and the easy addition of new tokens. 
The vocabulary size was set to 32,768, with around 400 tokens reserved for future expansion to maintain a multiple of 8.
We finally train the tokenizer on a subsample of our pre-training dataset that include a subsample of CulturaX and full French Wikipedia and HAL.

\subsection{Pre-Training Methodology}
The pre-training process for both CamemBERTv2 and CamemBERTav2 models was done in two stages. 
Initially, both models were trained with a sequence length of 512 tokens, which allowed for faster convergence during the early stages of training. 
In the second stage, the models were further pre-trained using a sequence length of 1024 tokens to fully capture long-range dependencies and improve performance on tasks requiring extensive context.
To create a pre-training dataset for the long sequence training, we further filter our pretraining corpus to have only long documents, while also including short sequences with a 5\% chance to ensure the model retains the ability to correctly handle shorter sequences.

Additionally, it was shown by \cite{antoun2023data} that models trained with MLM require multiple epochs of pre-training to achieve optimal accuracy, due to the Masked Language modeling objective only being able to propagate the loss from the masked tokens.
Hence, we train CamemBERTv2 for three epochs over our dataset.
We set the token masking rate to 40\%, which was found to be optimal by \citet{wettig-etal-2023-mask}.
In contrast, CamemBERTav2, being based on the more sample-efficient DeBERTaV3 pertaining methodology of replaced-token detection, reaches peak performance after just one epoch, making it significantly more efficient in terms of training time and computational resources.
Details about pre-training hyperparameters are available in Table~\ref{table:pre-train-hp}

\section{Experiments and Results}
\subsection{Downstream Evaluation}
\paragraph{General Domain}
To evaluate our models we consider a range of French downstream tasks and datasets, including Question Answering (QA) using FQuAD 1.0~\cite{2020arXiv200206071}, Part-Of-Speech (POS) tagging and Dependency Parsing on GSD~\cite{mcdonald-etal-2013-universal}, Rhapsodie~\cite{lacheret:halshs-01061368}, Sequoia~\cite{candito-seddah-2012-le,CANDITO14.494} in their UD v2.2 formats, and the French Social Media Bank~\cite{seddah-etal-2012-french}. 
We also assess Named Entity Recognition (NER) on the 2008 FTB version~\cite{abeille-etal-2000-building,candito-crabbe-2009-improving} with NER annotations by \citet{sagot-etal-2012-annotation}.
To assess text classification capabilities we evaluate our models on the FLUE benchmark~\cite{le2019flaubert}.
We re-used the same splits from~\citet{antoun2023data}, and performed hyper-parameter tuning on all models and datasets with 5 seed variations, except the dependency parsing and part-of-speech tasks where we only validate over 5 seeds using the same sets of parameters.

\paragraph{Domain Specific}
To assess the models on domain-specific tasks, we include the French subset of the pseudoanonymized dataset for radicalization detection with NER annotations~\cite{riabi-etal-2024-cloaked} which we refer to as Counter-NER.
For biomedical-domain datasets, we evaluate five distinct tasks: EMEA, MEDLINE, CAS1, CAS2, and E3C. EMEA and MEDLINE are part of the QUAERO corpus~\cite{grouin2014quaero}, where EMEA consists of drug leaflets and MEDLINE includes scientific article titles, both annotated with 10 semantic groups from the UMLS. 
CAS1 and CAS2 are based on the CAS corpus~\cite{grouin-etal-2019-clinical}, focusing on pathology and symptoms in the first subtask, while the second subtask involves extracting additional clinical information such as anatomy and treatment. 
Finally, E3C~\cite{Magnini2020TheEP} focuses on clinical cases from scientific articles, using fully annotated texts to identify clinical entities. For consistency, we adopt the dataset splits and hyper-parameters proposed by~\citet{touchent-de-la-clergerie-2024-camembert} for comparison with his model.

\subsection{General Domain Results}
For general domain tasks, the results show clear performance trends between models:
\paragraph{POS Tagging and Dependency Parsing:} As shown in Table~\ref{tab:pos_and_dp_results}, all models performed well on Universal POS (UPOS) tagging and dependency parsing, where the updated CamemBERTv2 and CamemBERTav2 model maintaining the previous models' scores.
These results indicate a possible saturation in the benchmark scores for current encoder-based transformer models.

\paragraph{Named Entity Recognition (NER):} In general domain NER, evaluated on the FTB dataset, CamemBERTaV2 outperformed all other models with an F1 score 93.4\% showing a significant improvement over the baseline CamemBERT model (89.97\%), while also improving over the MLM-trained CamemBERTv2 model.

\paragraph{Question Answering (QA):} For the FQuAD 1.0 dataset (Table~\ref{tab:qa}), CamemBERTav2 achieved the highest F1 score (83.04\%) and exact match (EM) score (64.29\%), outperforming the other models by a significant margin. The performance gap between CamemBERTv2 and CamemBERTav2 (80.39\% vs 83.04\%) suggests that the latter's enhanced pre-training loss and architecture yielded more robust representations for machine comprehension tasks in French.

\begin{table}[H]
    \centering
    {\footnotesize
        \begin{tabular}{lcc}
            \toprule
            Model               & F1             & EM             \\
            \midrule
            CamemBERT          & 80.98{\scriptsize$\pm$0.48} & 62.51{\scriptsize$\pm$0.54} \\
            CamemBERTa         & 81.15{\scriptsize$\pm$0.38} & 62.01{\scriptsize$\pm$0.45} \\
            \midrule
            CamemBERTv2        & 80.39{\scriptsize$\pm$0.36} & 61.35{\scriptsize$\pm$0.39} \\
            CamemBERTav2       & \textbf{83.04}{\scriptsize$\pm$0.19} & \textbf{64.29}{\scriptsize$\pm$0.31} \\            
            \bottomrule
        \end{tabular}
    }%
    \caption{Question Answering results on FQuAD 1.0.}
    \label{tab:qa}
\end{table}

\paragraph{Text Classification:} Table~\ref{tab:text_classification_results} presents text classification results across the CLS, PAWS-X, and XNLI tasks from the FLUE benchmark. CamemBERTav2 consistently outperformed other models, achieving top scores in all tasks, with the highest accuracy on the CLS task (95.63\%), PAWS-X (93.06\%), and XNLI (84.82\%). 
The massive increase in CamemBERTav2's XNLI scores compared to the previous CamemBERTa model shows that small transformer-based models, that use the sample-efficient RTD objective, can still benefit from increasing the unique token count during pretraining.

\begin{table}[h]

    \small\centering
\resizebox{\columnwidth}{!}{%
    \begin{tabular}{lccc}
        \toprule
        Model               & \textsc{CLS} & \textsc{PAWS-X} & \textsc{XNLI}  \\
        \midrule
        CamemBERT           & 94.62{\scriptsize$\pm$0.04}   &  91.36{\scriptsize$\pm$0.38} & 81.95{\scriptsize$\pm$0.51} \\
        CamemBERTa          & 94.92{\scriptsize$\pm$0.13}   & 91.67{\scriptsize$\pm$0.17}  & 82.00{\scriptsize$\pm$0.17} \\
        \midrule
        CamemBERTv2         & 95.07{\scriptsize$\pm$0.11}   &  92.00{\scriptsize$\pm$0.24} & 81.75{\scriptsize$\pm$0.62} \\
        CamemBERTav2        & \textbf{95.63}{\scriptsize$\pm$0.16}   & \textbf{93.06}{\scriptsize$\pm$0.45}  & \textbf{84.82}{\scriptsize$\pm$0.54} \\
        \bottomrule
    \end{tabular}%
    }
    \caption{Text classification results (Accuracy) on the FLUE benchmark.}
    \label{tab:text_classification_results}
\end{table}

\subsection{Domain Specific Results}
In the evaluation of domain-specific tasks, Table~\ref{tab:domain-specific-results-small}, particularly in the medical fields, both CamemBERTv2 and CamemBERTav2 exhibited strong performance. 
On medical NER tasks, the new models were able to achieve results comparable to domain-specific models, namely CamemBERT-bio, showcasing their ability to handle specialized terminologies and complex entity recognition. 
Notably, CamemBERTv2 and CamemBERTav2 significantly outperformed their predecessors across all tasks, largely due to the inclusion of scientific and medical articles in their updated pre-training datasets. 

In the radicalization NER task, which involves identifying sensitive and domain-specific entities, both models demonstrated large improvements.
CamemBERTav2 surpassed the original CamemBERT model by 2 percentage points, while CamemBERTv2 exceeded CamemBERT by over 3 points, further highlighting the enhancements made in these newer versions. 
These gains showcase the models' ability to generalize to challenging, niche domains with specialized vocabularies.

\begin{table}[H]
    \centering
    {\footnotesize
        \begin{tabular}{lcc}
            \toprule
            Model               & Medical-NER & Counter-NER \\
            \midrule
            CamemBERT          & 70.96{\scriptsize$\pm$0.13} & 84.18{\scriptsize$\pm$1.23}  \\
            CamemBERTa         & 71.86{\scriptsize$\pm$0.11} & 87.37{\scriptsize$\pm$0.73} \\
            CamemBERT-bio      & \textbf{73.96}{\scriptsize$\pm$0.12} & - \\
            \midrule
            CamemBERTv2        & 72.77{\scriptsize$\pm$0.11} & 87.46{\scriptsize$\pm$0.62} \\
            CamemBERTav2       & \textbf{73.98}{\scriptsize$\pm$0.11} & \textbf{89.53}{\scriptsize$\pm$0.73} \\            
            \bottomrule
        \end{tabular}
    }%
    \caption{Summary of NER F1 scores on the domain-specific downstream tasks. Full scores are available in Table~\ref{tab:domain-specific-results}.}
    \label{tab:domain-specific-results-small}
\end{table}

\subsection{Discussion}
The results from our experiments clearly demonstrate the significant advancements that CamemBERTv2 and CamemBERTav2 bring to French NLP tasks, both in general and domain-specific contexts. 
In the general domain tasks, CamemBERTav2 consistently outperformed its predecessors, showcasing the effectiveness of the DeBERTaV3 architecture and the RTD objective in handling both contextual and positional representations. 
The improvements seen in tasks such as NER, QA, and text classification are particularly noteworthy. 
For example, in the FQuAD 1.0 dataset, the large performance gap between CamemBERTv2 and CamemBERTav2 illustrates the robustness of the latter in understanding complex queries and extracting relevant answers. 
The enhanced tokenizer, with its improved handling of French-specific linguistic features and expanded vocabulary, also played a key role in these improvements.

Interestingly, while the models achieved high accuracy in POS tagging and dependency parsing tasks, the marginal gains over the original CamemBERT suggest that transformer-based models may be approaching performance ceilings on these specific benchmarks. 
This observation indicates that future progress in these areas might require new approaches, such as task-specific architectures or training methodologies, rather than further refinements to existing models.

In domain-specific tasks, the inclusion of scientific and medical articles in the pre-training dataset allowed both CamemBERTv2 and CamemBERTav2 to achieve strong results across specialized fields. 
Their ability to generalize to biomedical NER tasks, where they performed comparably to models specifically designed for the medical domain, shows the versatility of our updated models. 
The sizable improvements in the radicalization NER task also reflect the enhanced knowledge embedded in the new models, which is essential for identifying sensitive and rare entities within challenging domains.

These results affirm the value of continual model updates, particularly in addressing the issue of temporal concept drift. 
As language evolves and new terminologies emerge, updating models with more recent datasets and architectures becomes crucial for maintaining their relevance and utility in real-world applications. 
Our decision to update the tokenizer to better handle modern language elements like emojis and numerical data further reinforces this point, allowing the models to stay aligned with contemporary communication patterns.
\section{Conclusion}
In conclusion, the development of CamemBERTv2 and CamemBERTav2 marks a significant advancement in French language modeling, demonstrating improved performance across a variety of general and domain-specific NLP tasks. 
By leveraging larger and more recent datasets, alongside an updated tokenizer, these models have shown enhanced versatility and robustness, particularly in tasks like NER, QA, and text classification. 
However, the marginal improvements seen in certain tasks like POS tagging and dependency parsing suggest that these benchmarks may be nearing saturation for current transformer-based models.

Looking ahead, future work should not only focus on refining model architectures and training objectives but also prioritize updating datasets. 
Temporal concept drift is not solely a model issue—it is also a dataset issue. 
Many benchmarks currently in use do not reflect the latest linguistic distributions, which can exacerbate the performance gap between models trained on outdated versus modern data. 
Ensuring that datasets are regularly updated to include contemporary topics, terminologies, and language use is essential for keeping models relevant and maximizing their real-world applicability. 
Such efforts will ensure that both models and benchmarks evolve together, addressing temporal drift more effectively and pushing the boundaries of what these systems can achieve.

\section*{Acknowledgements}

This work was partly funded by Benoît Sagot's chair in the PRAIRIE institute funded by the French national reseach agency (ANR as part of the ``Investissements d’avenir'' programme under the reference \mbox{ANR-19-P3IA-0001}. 
This work  also received funding from the European Union’s Horizon 2020
research and innovation program under grant agreement No. 101021607. 
The authors are grateful to the OPAL infrastructure from Université Côte d'Azur for providing resources and support.
This work was also granted access by GENCI to the HPC resources of IDRIS under the
allocation 2024-GC011015610.
Finally, part of this work was funded by the DINUM through the \href{https://alliance.numerique.gouv.fr/les-produits-incubés/camembert2_0/}{AllIAnce program}.

We would also like to thank Nathan Godey, and Arij Riabi for the productive
discussions.

\bibliography{custom.bib}

\newpage

\appendix

\section{Full Results}
\label{sec:appendix}

\subsection{Full Domain Specific NER Results}

\begin{table}[H]
\small\centering
\resizebox{0.9\columnwidth}{!}{
\begin{tabular}{llc}
\toprule
Dataset & Model &          F1    \\
\midrule
\multirow{6}{*}{CAS1}       & CamemBERT &  70.72{\scriptsize$\pm$1.47} \\
                            & CamemBERTa &  71.96{\scriptsize$\pm$1.38} \\ \cline{2-3} \\ [-2ex]
                            & Dr-BERT &  62.76{\scriptsize$\pm$1.55} \\
                            & CamemBERT-Bio &  72.28{\scriptsize$\pm$1.46} \\ \cline{2-3} \\ [-2ex]
                            & CamemBERTv2 &  71.18{\scriptsize$\pm$1.62} \\
                            & CamemBERTav2 &  \textbf{72.87}{\scriptsize$\pm$2.29} \\
\midrule
\multirow{6}{*}{CAS2}       & CamemBERT &  78.43{\scriptsize$\pm$1.78} \\
                            & CamemBERTa &  79.06{\scriptsize$\pm$0.68} \\ \cline{2-3} \\ [-2ex]
                            & Dr-BERT &  76.43{\scriptsize$\pm$0.49} \\
                            & CamemBERT-Bio &  \textbf{82.50}{\scriptsize$\pm$0.56} \\ \cline{2-3} \\ [-2ex]
                            & CamemBERTv2 &  81.87{\scriptsize$\pm$0.58} \\
                            & CamemBERTav2 &  81.85{\scriptsize$\pm$0.49} \\
\midrule
\multirow{6}{*}{E3C}        & CamemBERT &  67.01{\scriptsize$\pm$2.13} \\
                            & CamemBERTa &  67.01{\scriptsize$\pm$1.85} \\ \cline{2-3} \\ [-2ex]
                            & Dr-BERT &  56.99{\scriptsize$\pm$2.40} \\
                            & CamemBERT-Bio &  69.87{\scriptsize$\pm$1.21} \\ \cline{2-3} \\ [-2ex]
                            & CamemBERTv2 &  69.27{\scriptsize$\pm$0.90} \\
                            & CamemBERTav2 &  \textbf{70.12}{\scriptsize$\pm$0.87} \\
\midrule
\multirow{6}{*}{EMEA}       & CamemBERT &  73.53{\scriptsize$\pm$2.04} \\
                            & CamemBERTa &  75.99{\scriptsize$\pm$0.51} \\ \cline{2-3} \\ [-2ex]
                            & Dr-BERT &  71.33{\scriptsize$\pm$0.84} \\
                            & CamemBERT-Bio &  76.96{\scriptsize$\pm$2.00} \\ \cline{2-3} \\ [-2ex]
                            & CamemBERTv2 &  76.30{\scriptsize$\pm$1.00} \\
                            & CamemBERTav2 &  \textbf{77.28}{\scriptsize$\pm$0.57} \\
\midrule
\multirow{6}{*}{MEDLINE}    & CamemBERT &  65.11{\scriptsize$\pm$0.56} \\
                            & CamemBERTa &  65.33{\scriptsize$\pm$0.30} \\ \cline{2-3} \\ [-2ex]
                            & Dr-BERT &  58.90{\scriptsize$\pm$0.51} \\ 
                            & CamemBERT-Bio &  \textbf{68.21}{\scriptsize$\pm$0.91} \\ \cline{2-3} \\ [-2ex]
                            & CamemBERTv2 &  65.26{\scriptsize$\pm$0.33} \\
                            & CamemBERTav2 &  67.77{\scriptsize$\pm$0.44} \\
\midrule
\multirow{4}{*}{Counter-NER} & CamemBERT &  84.18{\scriptsize$\pm$1.23} \\
                             & CamemBERTa &  87.37{\scriptsize$\pm$0.73} \\ \cline{2-3} \\ [-2ex]
                             & CamemBERTv2 &  87.46{\scriptsize$\pm$0.62} \\
                             & CamemBERTav2 &  \textbf{89.53}{\scriptsize$\pm$0.73} \\
\bottomrule
\end{tabular}
}
\caption{NER F1 scores on the domain-specific downstream tasks.}
\label{tab:domain-specific-results}
\end{table}

\clearpage
\onecolumn

\section{Hyper-parameters}
\subsection{Pre-training Hyper-parameters}

\begin{table*}[htb!]
    \centering
    \scalebox{1}{
    \begin{tabular}{|@{\hskip3pt}l@{\hskip2pt}|@{\hskip2pt} c@{\hskip2pt}|@{\hskip2pt} c@{\hskip2pt}|}
        \toprule
          Hyper-parameter & {CamemBERTav2}\textsubscript{base} &  {CamemBERTv2}\textsubscript{base} \\
        \midrule
        Number of Layers & 12 & 12 \\
        Hidden size & 768 & 768 \\
        Generator Hidden size & 256 & - \\
        FNN inner Hidden size & 3072 & 3072 \\
        Attention Heads  & 12  & 12\\
        Attention Head size & 64 & 64\\
        Dropout & 0.1& 0.1 \\
        Warmup Steps (p1/p2) & 10k/1k & 10k/1k \\
        Learning Rates (p1/p2) & 7e-4/3e-4 & 7e-4/3e-4 \\
        End Learning Rates (p1/p2) & 1e-5& 1e-5 \\
        Batch Size& 8k& 8k \\
        Weight Decay & 0.01& 0.01 \\
        Max Steps (p1/p2) & 91k/17k & 273k/17k \\
        Learning Rate Decay&  Polynomial p=0.5 &  Polynomial p=0.5 \\
        Adam $\epsilon$ & 1e-6&1e-6 \\
        Adam $\beta_1$ & 0.878& 0.878 \\
        Adam $\beta_2$ & 0.974& 0.974 \\
        Gradient Clipping& 1.0& 1.0 \\ 
        Masking Probability & 20\% & 40\% \\
        Seq. Length (p1/p2) & 512/1024& 512/1024 \\
        Precision & BF16 & BF16 \\
        \bottomrule
        \end{tabular}
        }
    \caption{Hyper-parameters for pre-training {CamemBERTa and CamemBERT 2.0}. }
    \label{table:pre-train-hp}
\end{table*}

\subsection{Fine-Tuning Hyper-parameters}

\begin{table*}[h]
\centering

\scalebox{0.88}{
\begin{tabular}{|c|c|c|c|c|c|c|}
\hline
\textbf{Task}         & \textbf{Learning Rate} & \textbf{LR Sch.} & \textbf{Epochs} & \textbf{Max Len.} & \textbf{Batch Size} & \textbf{Warmup}\\ \hline
\textbf{FQuAD}        & \{3, 5, 7\}e-5 & cosine            & 6            & 1024  & \{32,64\} & \{0,0.1\} \\ \hline
\textbf{CLS}          & \{3, 5, 7\}e-5 & \begin{tabular}[c]{@{}c@{}} cosine \\ linear\end{tabular} & 6            & 1024  & \{32,64\} & 0         \\ \hline
\textbf{PAWS-X}       & \{3, 5, 7\}e-5 & \begin{tabular}[c]{@{}c@{}} cosine \\ linear\end{tabular} & 6            & 148   & \{32,64\} & 0         \\ \hline
\textbf{FTB NER}      & \{3, 5, 7\}e-5 & \begin{tabular}[c]{@{}c@{}} cosine \\ linear\end{tabular} & 8            & 192   & \{16,32\} & \{0,0.1\} \\ \hline
\textbf{XNLI}         & \{3, 5, 7\}e-5 & cosine            & 10           & 160   & 32        & 0.1       \\ \hline
\textbf{POS}          & 3e-05                                                                       & linear            & 64           & 1024  & 8         & 100 steps \\ \hline
\textbf{Dep. Pars.}   & 3e-05                                                                      & linear            & 64           & 1024  & 8         & 100 steps \\ \hline
\textbf{Counter-NER}  & \{3, 5, 7\}e-5 & \begin{tabular}[c]{@{}c@{}} cosine \\ linear\end{tabular} & 8            & 512   & \{16,32\} & \{0,0.1\} \\ \hline
\textbf{Med-NER} & 5e-5 & linear & 3 & 20 & 8 & 0.224 \\ \hline
\end{tabular}
}
\caption{Hyperparameter Search During Fine-tuning of CamemBERTv2. All models were trained with FP32}
\label{tab:hyperparameters}
\end{table*}

\end{document}